\documentclass[journal]{IEEEtran}
\usepackage{graphicx}
\usepackage{cite}
\usepackage{picinpar}
\usepackage{url}
\usepackage{pifont}
\usepackage[latin1]{inputenc}
\usepackage{colortbl}
\usepackage{soul}
\soulregister\cite7
\soulregister\ref7
\usepackage{multirow}
\usepackage{pifont}
\usepackage{color}
\usepackage{alltt}
\usepackage[hidelinks]{hyperref}
\usepackage{enumerate}
\usepackage{siunitx}
\usepackage{breakurl}
\usepackage{epstopdf}
\usepackage{pbox}
\usepackage{amsmath,amssymb,amsfonts}
\usepackage{textcomp}
\usepackage{wrapfig}
\usepackage{empheq}
\usepackage{array}
\usepackage{makecell}
\usepackage{xcolor}
\usepackage{booktabs}
\usepackage{threeparttable}
\usepackage{adjustbox}
\usepackage{subfigure}
\usepackage{verbatim}
\usepackage{gensymb}
\usepackage[misc,geometry]{ifsym}
\graphicspath{{figures/}}
\definecolor{ccr}{RGB}{10,50,200}
\usepackage{hyperref}
\hypersetup{hypertex=true,
	colorlinks=true,
	linkcolor=ccr,
	anchorcolor=ccr,
	citecolor=ccr}
\usepackage{multicol}
\usepackage[algoruled,vlined,linesnumbered]{algorithm2e}
\usepackage{caption}
\usepackage{lipsum} 
\usepackage{lineno}

\begin{document}
	
	\title{ForeRobo: Unlocking Infinite Simulation Data for 3D Goal-driven Robotic Manipulation}
	
	\author{
		\vskip 1em
		
		Dexin Wang, Faliang Chang, Chunsheng Liu
		
		\thanks{
			All authors are with the School of Control Science and Engineering, Shandong University, Ji'nan, Shandong 250061, China (e-mail: flchang@sdu.edu.cn, liuchunsheng@sdu.edu.cn). (Corresponding author: Faliang Chang, Chunsheng Liu)
		}
	}
	
	\maketitle
	
	\begin{abstract}
		Efficiently leveraging simulation to acquire advanced manipulation skills is both challenging and highly significant.
        We introduce \textit{ForeRobo}, a generative robotic agent that utilizes generative simulations to autonomously acquire manipulation skills driven by envisioned goal states.
        Instead of directly learning low-level policies, we advocate integrating generative paradigms with classical control.
        Our approach equips a robotic agent with a self-guided \textit{propose-generate-learn-actuate} cycle. 
        The agent first proposes the skills to be acquired and constructs the corresponding simulation environments; it then configures objects into appropriate arrangements to generate skill-consistent goal states (\textit{ForeGen}). 
        Subsequently, the virtually infinite data produced by ForeGen are used to train the proposed state generation model (\textit{ForeFormer}), which establishes point-wise correspondences by predicting the 3D goal position of every point in the current state, based on the scene state and task instructions. 
        Finally, classical control algorithms are employed to drive the robot in real-world environments to execute actions based on the envisioned goal states.
        Compared with end-to-end policy learning methods, ForeFormer offers superior interpretability and execution efficiency. 
        We train and benchmark ForeFormer across a variety of rigid-body and articulated-object manipulation tasks, and observe an average improvement of 56.32\% over the state-of-the-art state generation models, demonstrating strong generality across different manipulation patterns. 
        Moreover, in real-world evaluations involving more than 20 robotic tasks, ForeRobo achieves zero-shot sim-to-real transfer and exhibits remarkable generalization capabilities, attaining an average success rate of 79.28\%.

	\end{abstract}
	

	\section{Introduction}
    Simulation environments, due to their efficiency and convenience, have become a crucial means for robotic skill acquisition \cite{mandlekar2022matters} \cite{liu2023libero}. 
    However, simulation inaccuracies in physical dynamics and material properties constrain the effective transfer of trained policies to real-world scenarios.

    Many researchers have explored various approaches to acquiring robotic skills in simulation, aiming to strike a balance between real-world transferability and policy completeness.
    Advocates of directly learning low-level policies employ reinforcement learning \cite{luo2025precise} \cite{cui2024task} or imitation learning \cite{wang2025hierarchical} \cite{ma2024hierarchical} to map from 2D/3D observations to actions, but their real-world transferability is constrained by visual discrepancies and kinematic gaps.
    Advocates of trajectory-prediction-based planning attempt to construct world models to capture environment state transitions \cite{jain2025smooth} \cite{zhang2025flowpolicy}, but these approaches tend to suffer from error accumulation.
    Advocates of decoupling object motion from robot actions predict goal object poses \cite{pan2023tax} \cite{wang2024learning} or point cloud flows \cite{guo2025flowdreamer} \cite{seita2023toolflownet} to drive robotic manipulation, achieving impressive multi-entity generalization \cite{zhi20253dflowaction}, though current methods still lack task-level generalization.
    Consequently, how to leverage simulation to acquire policies with zero-shot sim-to-real transferability, low error, and task-level generalization remains an open and pressing challenge.

    In light of these challenges, we propose a novel paradigm characterized by three key aspects: (1) exploiting the low-error components of simulators in specific physical attributes to train models and enable zero-shot sim-to-real transfer; (2) decoupling object motion from robot actions to enhance multi-entity generalization; and (3) scaling up simulation-based robot learning to achieve task-level generalization.

    In this paper, as an initial realization of this paradigm, we introduce \textit{ForeRobo}, a robotic agent capable of autonomously acquiring new skills through a self-guided propose-generate-learn-actuate cycle.
    First, it leverages LLM to self-propose tasks and configure environments, including the required assets and initial states.
    Next, it receives a single human-provided goal-state demonstration for each skill and employs the proposed cross-instance proximity contact alignment technique (\textit{CPCA}) to generate robot-agnostic, skill-consistent goal states across all scenarios. We refer to this process as \textit{ForeGen}.
    Next, it trains the proposed 3D goal-state generation model \textit{ForeFormer} entirely on simulated data, which establishes point-wise correspondences by predicting each point's goal position conditioned on the scene and task instructions; finally, guided by the foresighted goal states predicted by ForeFormer, ForeRobo employs classical control methods to drive robots in real-world settings to accomplish manipulation tasks.

    The proposed ForeRobo offers the following advantages:
    \begin{itemize}
        \item 
        \textbf{Strong foresight.}
        It leverages predicted goal states to guide global planning, enhancing trajectory smoothness, execution efficiency, and manipulation robustness, while also allowing the robot's intentions to be anticipated, thereby improving system interpretability.
    
        \item 
        \textbf{Zero-shot sim-to-real.}
        By representing states with low-simulation-error 3D point clouds and integrating obstacle-aware planning in real-world environments, ForeRobo can be directly deployed without any fine-tuning.
    
        \item 
        \textbf{Multi-entity generalization.}
        Decoupling object motion from robot actions enables the use of robot-specific motion planning algorithms across different platforms.

        \item 
        \textbf{Manipulation-ready goal states.}
        Compared with goal states represented by images or optical flow \cite{blackzero}, \textit{ForeFormer} predicts the target position of each point on the object, enabling convenient extraction of both global and local structured object motions (e.g., pose transformations), which can directly drive robotic manipulation.
    
        \item 
        \textbf{Rapid scalability.}
        The automatic generation of goal states fully exploits the potential of virtually infinite simulation data, leading to substantial task-level generalization.
    \end{itemize}

    \begin{figure*}[tp]
		\centering
		{\includegraphics[scale=1]{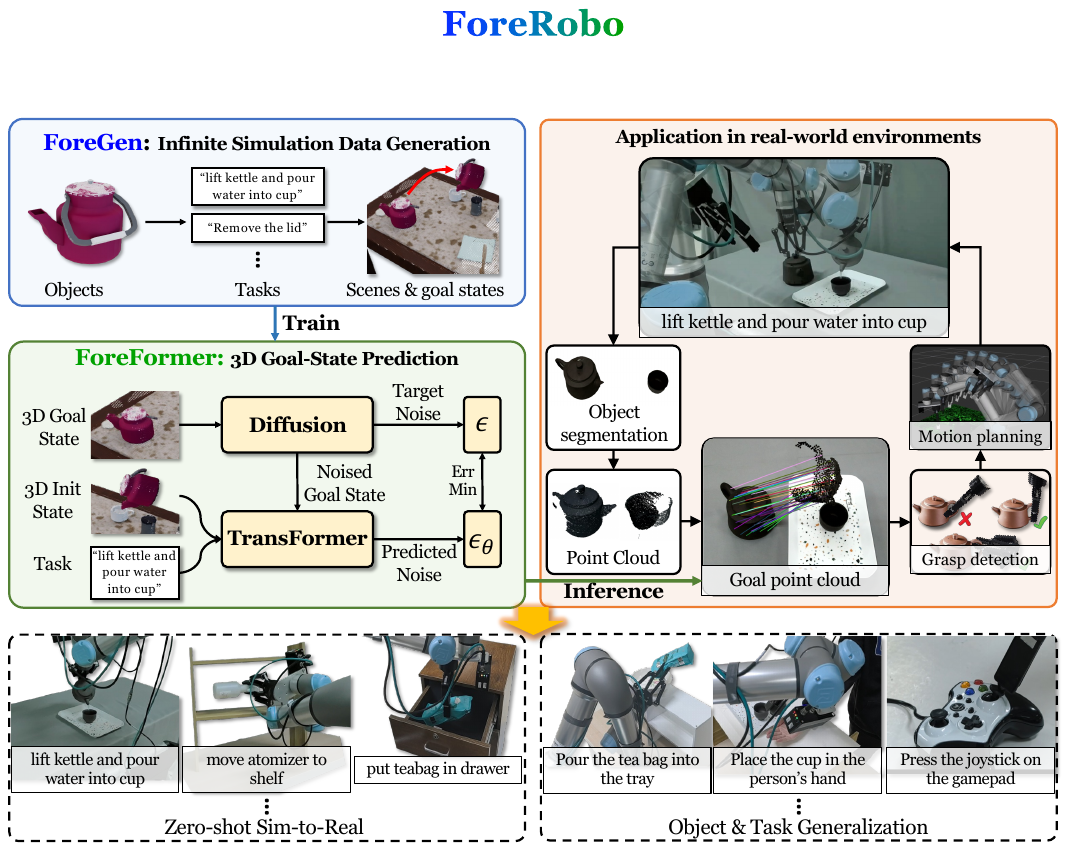}}
		\caption{
			\textbf{Overview of ForeRobo}. 
            ForeRobo primarily consists of a data generation component, \textit{ForeGen}, and a state prediction model, \textit{ForeFormer}. 
            The top-right part of the figure demonstrates how ForeFormer, trained exclusively on simulation data, enables the robot arm to accomplish manipulation tasks in real-world environments. The process mainly involves five steps: (1) task-relevant object segmentation; (2) object point cloud acquisition; (3) goal state prediction using ForeFormer; (4) grasp detection; and (5) robot motion planning. Detailed descriptions of these steps are provided in the Appendix.
            The bottom part of the figure illustrates that ForeFormer, trained entirely on simulation data, can be zero-shot transferred to real-world environments and generalize across diverse objects and manipulation tasks.
		}
		\label{fig_ForeRobo}
	\end{figure*}

    Our experiments demonstrate that the first version of ForeRobo provides a diverse set of skills covering both rigid and articulated objects, including pick-place, pour, open, close, push and so on (see Figure~\ref{fig_ForeRobo}).
    We conducted comprehensive ablation studies to identify the optimal model configuration, including both model architecture and loss function design.
    Additionally, we evaluated the performance differences between our method and recent goal state generation models across a variety of simulated and real-world tasks.
    On 10 simulation tasks, ForeFormer achieved an average performance improvement of 47.14\% over recent state generation models.
    On 21 real-world tasks, ForeFormer, trained solely on simulation data generated by ForeGen, reaches an average success rate of 79.28\%, representing an improvement of 65.50\% over recent state generation models, and demonstrates strong sim-to-real transferability and task-level generalization.
    Overall, the results indicate that ForeRobo outperforms current mainstream methods in terms of generalization, execution efficiency, interpretability, and controllability.

    \begin{figure*}[tp]
		\centering
		{\includegraphics[scale=0.45]{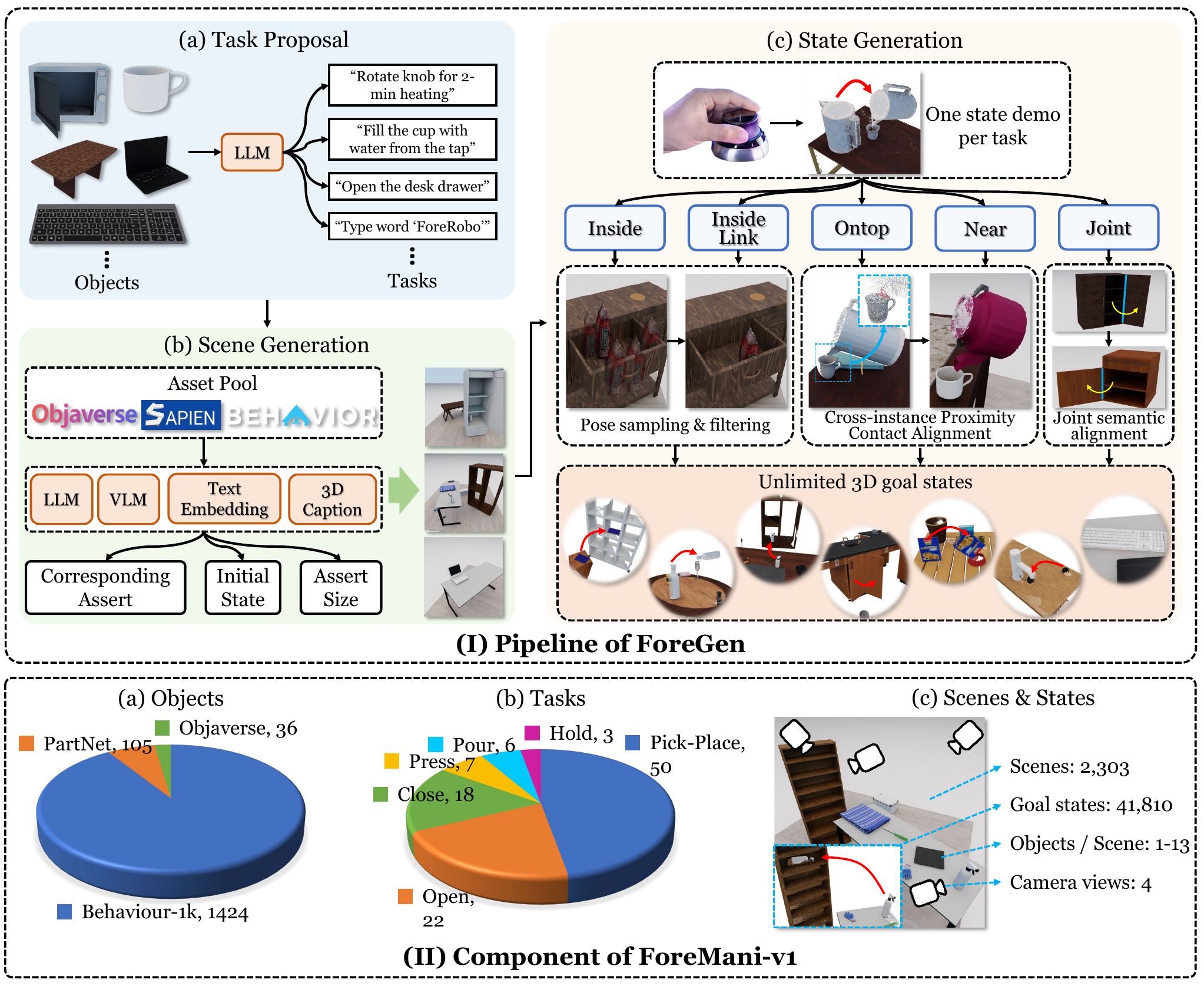}}
		\caption{
			\textbf{Overview of ForeGen and ForeMani-v1}.
            ForeGen consists of the following stages: a) task proposal, b) scene generation, and c) state generation.
            ForeMani-v1 encompasses 1,536 objects and 106 tasks, with each task containing an average of 21 scenarios and 394 goal states.
		}
		\label{fig_ForeGen}
	\end{figure*}

    \section{ForeGen}
    ForeGen is an automated pipeline that leverages state-of-the-art foundational models to systematically generate skills, scenes, and foresight states.
    As an initial realization of our proposed paradigm, ForeGen-v1 focuses on manipulation skills involving rigid and articulated objects.
    We illustrate the complete pipeline in Figure~\ref{fig_ForeGen}, which consists of three integral stages: Task Proposal, Scene Generation, and State Generation.
    We detail each stage in the following.

    \subsection{Task Proposal}
    ForeGen begins by proposing meaningful tasks suitable for robotic learning.
    It first randomly samples a rigid or articulated object from a predefined asset pool, and then queries a state-of-the-art large language model (LLM), leveraging commonsense knowledge and an understanding of object affordances, to generate task proposals involving the sampled object.

    We use GPT-4 \cite{achiam2023gpt} as a queryable large language model (LLM) and adopt Behaviour-1K \cite{li2024behavior} and PartNetMobility \cite{mo2019partnet} as asset pools for sampling objects.
    The prompt fed to GPT-4 contains the following information: 1) the category of the sampled object; 2) a description of the object's appearance, obtained by analyzing multi-view images with Cap3D \cite{luo2023scalable} and subsequently merged via GPT-4; 3) semantic annotations of all links, if applicable. 
    GPT-4 generates a set of robot-executable tasks based on object affordances, with each task including: 1) a task description; 2) the link involved in the task, if applicable; 3) additional objects required to perform the task.
    By repeatedly querying with different sampled objects, ForeGen can generate a diverse range of manipulation tasks.

    \subsection{Scene Generation}
    For each proposed task, ForeGen subsequently configures each object to generate the corresponding scene, and rapidly produces diverse scene augmentation instances by varying the assets and states of objects within the scene.
    Each object is configured with the following elements: 1) the corresponding asset; 2) initial state; 3) asset size.
    We provide further details in the following.

    \textbf{Corresponding assert}
    For the additional objects required by the GPT-4-proposed tasks, ForeGen selects suitable models from the asset pools for instantiation.
    We adopt Behaviour-1K \cite{li2024behavior}, PartNetMobility \cite{mo2019partnet}, and Objaverse \cite{deitke2023objaverse} as the asset pools, containing a total of 1,565 assets. 
    Notably, Objaverse only includes non-articulated rigid objects.
    ForeGen first encodes the descriptions of required objects and all asset descriptions using SentenceTransformer \cite{reimers2019sentence} and computes their cosine similarity. It then queries GPT-4 to select the most suitable model from the top-$10$ most similar assets.
    During scene augmentation, ForeGen samples assets from the same-category asset sets within the corresponding datasets to instantiate the objects.
    Compared to repeatedly selecting assets based on similarity, this approach accelerates the construction of augmented scenes while ensuring semantic consistency of objects across different scenes for the same task.

    \textbf{Initial State}
    To enable effective skill acquisition, objects in each task must be initialized with appropriate spatial relations and joint configurations. 
    We predefined three types of spatial relations and employed GPT-4 to suggest suitable relations for each object: OnTop, Inside, and Inside link.
    For instance, in the task "make apple juice``, a plausible initial scene configuration is as follows: the apple is \textit{inside} the storage box, the storage box is \textit{inside} the table drawer \textit{link}, and the juicer is placed \textit{on top} of the table.
    To instantiate these relations, ForeGen samples candidate object poses in simulation and filters them to satisfy the predefined constraints while avoiding collisions. 
    For example, in the case of the "$A$ \textit{Inside} $B$``, the axis-aligned bounding box (AABB) of $A$ must be entirely contained within the AABB of $B$, while also satisfying one of the following conditions: 1) $A$ is located between the top and bottom surfaces of $B$; 2) $A$ is located between the four horizontal side surfaces of $B$.
    The initial joint angles of articulated objects are determined by querying GPT-4 for their normalized values.
    During scene augmentation, the spatial relations among objects remain fixed and are instantiated through the same sampling procedure described above.
    The joint angles of articulated objects are determined by aligning them with the link semantics of the corresponding objects in the initial scene.
   
    \textbf{Assert size}
    Appropriate asset sizing is essential for maintaining valid spatial relationships.
    Considering that the asset sizes in Behaviour-1K are aligned with real-world scales, ForeGen directly uses the original sizes for assets within Behaviour-1K. For assets outside Behaviour-1K, it searches for the most semantically similar asset within Behaviour-1K and reuses its size.
    During scene augmentation, objects outside Behaviour-1K retain the sizes assigned to the corresponding objects in the initial scene.

    \subsection{State Generation}
    Given the non-negligible kinematic simulation gaps in current simulators, to achieve zero-shot sim-to-real skill transfer and rapid, scalable skill acquisition, we advocate learning robot-action-independent goal state generation strategies from simulation, which are then used to drive robot manipulation in real-world settings.
    We predefine five types of goal states to cover most common rigid and articulated object manipulation tasks: 1) OnTop; 2) Inside; 3) Inside link; 4) Near; 5) Joint.
    Here, \textit{Near} indicates that objects are adjacent but not in contact. For example, in the task of ``pouring water from a kettle into a cup", the goal state is ``kettle Near cup".
    For each task, ForeGen uses a single human-demonstrated goal state in the initially generated scene as a reference and generates goal states in augmented scenes with different assets and layouts, resulting in multiple initial-goal state pairs for each task.
    For the \textit{OnTop} and \textit{Near} states, we design a Cross-instance Proximity Contact Alignment (CPCA) technique for state generation, with detailed description provided below.
    The \textit{Inside} state is generated through pose sampling with conditional filtering, following the same procedure used to establish \textit{Inside} spatial relationships.
    The \textit{Joint} states of articulated objects are determined by aligning their link semantics with those of demonstration objects from the same category.

    \begin{figure}[tp]
		\centering
		{\includegraphics[scale=0.37]{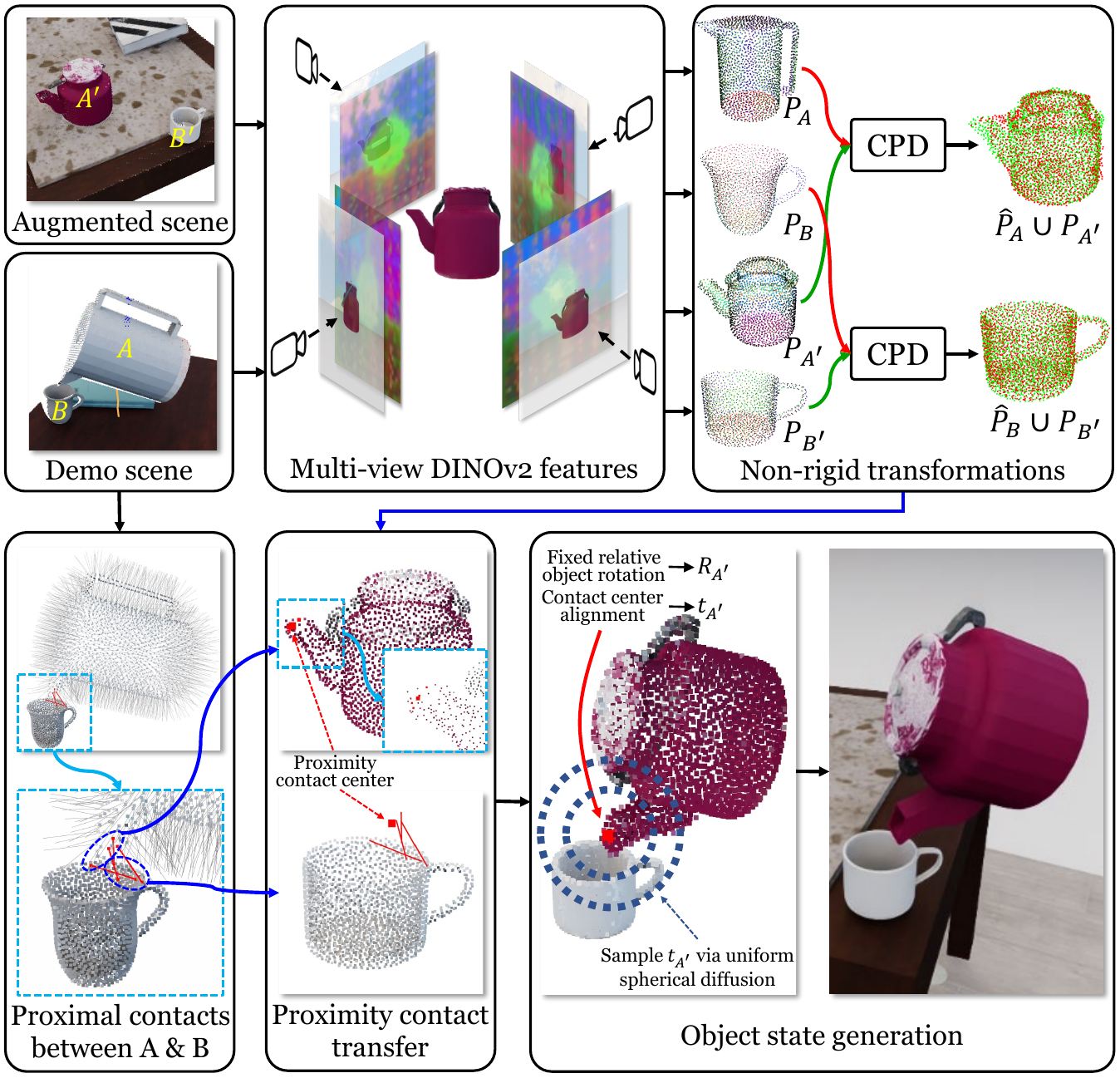}}
		\caption{
			\textbf{Pipeline of Cross-instance Proximity Contact Alignment (CPCA)}. 
            Given the demonstrated goal state in a task's demonstration scene, CPCA can generate the corresponding goal states for all augmented scenes of that task.
		}
		\label{fig_CPCA}
	\end{figure}

    During demonstration collection, we use a teleoperation device (i.e., spacemouse) to directly move objects or joints to the goal state, while recording object $B$ that is in close proximity to the manipulated object $A$.
    Given a pair of proximate objects $A$ and $B$ in the demonstration scene, and their corresponding objects $A'$ and $B'$ of the same category but different assets in an augmented scene, the CPCA pipeline consists of three stages: 1) object point cloud registration, 2) proximity contact transfer, 3) object state generation. 
    This process is illustrated in Figure~\ref{fig_CPCA}.
    We explain details in the following.
    
    \textbf{Object point cloud registration}
    CPCA first employs the DINOv2 model \cite{oquab2024dinov2} to extract semantically consistent multi-view visual features for each object, which are then embedded into the corresponding object point cloud to obtain a point cloud representation $P \in \mathbb{R} ^ {N \times (3+d)}$ enriched with semantic features.
    Subsequently, CPCA applies the Coherent Point Drift (CPD) algorithm \cite{myronenko2010point} to perform non-rigid transformations on the point clouds $P_A$ and $P_B$ of objects $A$ and $B$, aligning them respectively with $P_{A'}$ and $P_{B'}$:
    \begin{align}
		\hat{p_i} = p_i + u(p_i), ~~ 
        \hat{P} = \{ \hat{p_i} \}_{i=1}^N .
		\label{eq_1}
	\end{align}
    where, $p_i \in \mathbb{R}^3$ denotes a point in the point cloud $P$, $u(p_i) \in \mathbb{R}^3$ represents the displacement field applied to $p_i$, and $\hat{P}$ denotes the object point cloud after the non-rigid transformation.
    Benefiting from the semantic consistency of DINOv2 features across objects of the same category and their fine-grained discriminative ability over different object parts, CPCA enables accurate correspondence establishment between different regions of objects within the same class.
    Finally, for each point $\hat{p_i}$ in $\hat{P_A}$  and $\hat{P_B}$, a nearest-neighbor search is performed in $P_{A'}$ and $P_{B'}$ to identify the corresponding points, resulting in a set of point-wise correspondences:
    \begin{align}
		j(i) = \mathop{\arg\min}\limits_{j \in \{ 1,...,N' \}} \Vert \hat{p_i} - q_j \Vert,
        ~~
        \mathcal{C} = \{ (\hat{p_i}, q_{j(i)}) \}_{i=1}^N.
		\label{eq_2}
	\end{align}
    where, $N$ denotes the number of points in $P_A$ or $P_B$, and $N'$ denotes the number of points in $P_{A'}$ or $P_{B'}$.
    
    \textbf{Proximity contact transfer}
    The core idea of CPCA is based on the assumption that proximate contact patterns are consistent across different objects under the same state. By preserving the contact patterns of objects in the augmented scene consistent with those in the demonstration scene, CPCA generates the corresponding target states.
    For example, in the task of "pouring water from a kettle into a cup``, the lower surface of the kettle's spout maintains proximal contact with the inner wall of the cup across different scenes.
    CPCA detects the proximal contacts between objects $A$ and $B$ via ray casting, including their respective contact points, contact normal directions, and contact distances within a threshold. 
    Then, CPCA uses the point-wise correspondences between objects $A$ and $A'$ (see Eq.~\ref{eq_2}) to map the contact points into the coordinate frame of $A'$, denoted as $C_{A'} \in \mathbb{R} ^ {M \times 3}$, while preserving the contact directions $D_{A'} \in \mathbb{R} ^ {M \times 3}$ and contact distances $L_{A'} \in \mathbb{R} ^ {M \times 1}$ relative to the object. 
    The same procedure is applied to object $B'$.

    \textbf{Object state generation}
    To preserve the consistency of proximal contact patterns, CPCA keeps object $B'$ fixed and computes the rotation $R_{A'}$ and translation $t_{A'}$ of object $B'$ based on the invariance of the relative rotation between $A'$ and $B'$ and the alignment of proximal contact centers, as expressed in the following equation:
    \begin{align}
		R_A^{-1} R_B = R_{A'}^{-1} R_{B'}
        \\
        \sum_{i=1}^{M} (R_{B'} C_{B',i} + t_{B'} + R_{B'} D_{B',i} L_{B',i}) = \sum_{i=1}^{M} (R_{A'} C_{A',i} + t_{A'}).
		\label{eq_3}
	\end{align}
    where $M$ denotes the number of contact points. 
    By rearranging terms, we obtain:
    \begin{align}
		R_{A'} = R_{B'} R_B^{-1} R_A
        \\
        t_{A'} = \frac{\sum_{i=1}^{M} (R_{B'} C_{B',i} + t_{B'} + R_{B'} D_{B',i} L_{B',i} - R_{A'} C_{A',i})}{M}
		\label{eq_4}
	\end{align}
    To prevent collisions at the computed pose of object $A'$, coordinates are sampled outward uniformly on a sphere centered at $t_{A'}$, and the first collision-free pose is selected via simulation.
    This approach allows CPCA to maintain the spatial layout of the generated states consistent with the demonstration, thereby avoiding state deviations or semantic inconsistencies caused by distributional shifts.
    It should be emphasized that, although we only utilize CPCA to generate goal states in this work, the method can in principle generate any state involving proximal contacts, and by adjusting the contact distances, it can be extended to non-proximal contact states.
    We plan to further expand its range of applications in future work.

    By iteratively performing the above process, ForeGen can efficiently generate multiple tasks, scenes, and corresponding goal states.
    Moreover, since the proposed ForeRobo drives real-robot operations based on goal states, it avoids the sim-to-real gap associated with low-level policy transfer, thereby better leveraging the potential of infinite simulation data.
    Leveraging ForeGen, we develop ForeMani-v1, the first scalable manipulation dataset containing multi-task 3D goal states.
    ForeMani-v1 comprises 1,536 distinct object assets, spanning six categories of tasks: pick-place, open, close, press, pour, and hold, with a total of 106 task instances. 
    The dataset includes 2,303 simulation scenes and 41,810 3D goal states, averaging 21 scenes and 394 goal states per task.
    By iteratively executing the ForeGen pipeline, ForeMani-v1 can be easily extended.

    \section{ForeFormer}
    ForeFormer is a model that predicts 3D target states based on the initial scene state and task instructions, which are then used to drive robotic actions.
    We next provide a detailed description of the architecture of ForeFormer and its strategy for robot motion planning.

    \begin{figure}[tp]
		\centering
		{\includegraphics[scale=0.9]{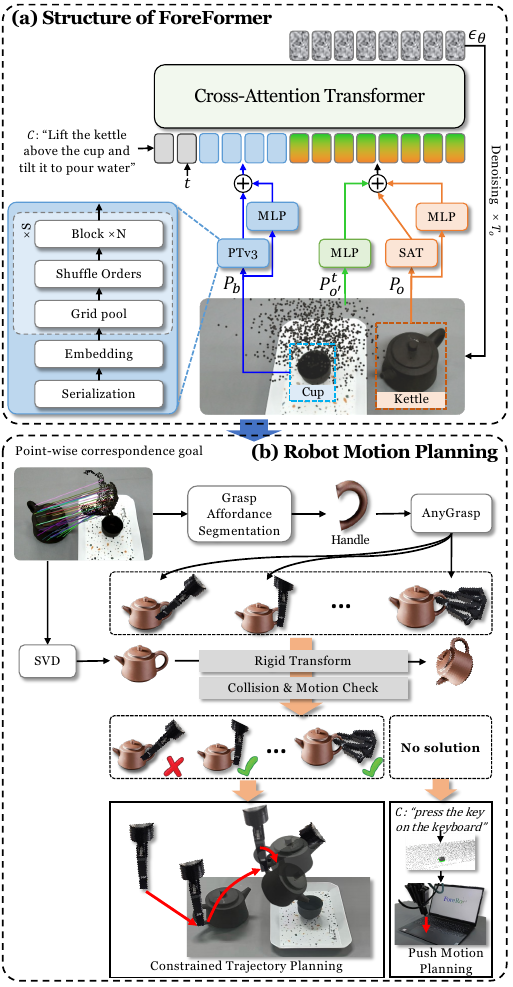}}
		\caption{
			\textbf{Overview of ForeFormer and robot motion planning}. 
            PTv3 refers to PointTransformer-v3, and SAT refers to the Self-Attention Transformer. 
            In the task "Lift the kettle above the cup and tilt it to pour water" depicted in the figure, the kettle is the object to be manipulated, which is embedded into the network via the object point cloud encoder (comprising SAT and two MLPs). The cup, being the task-relevant object, is embedded into the network through the background point cloud encoder (which includes PTv3 and an MLP).
		}
		\label{fig_ForeFormer}
	\end{figure}

    \subsection{ForeFormer Structure}
    ForeFormer takes the initial 3D scene state and task instructions as input and predicts the desired 3D goal state as output.
    Compared with directly outputting low-level actions or anticipated state trajectories, we advocate using the goal state to drive robotic operations, achieving a balance between reducing the sim-to-real gap and lowering learning complexity, while leveraging 3D states to enhance the robot's spatial perception capabilities.

    We formulate ForeFormer as Conditional Denoising Diffusion Probabilistic Models (CDDPMs) \cite{ho2020denoising} to capture multi-modal goal state distributions. 
    The 3D scene state is represented by a static background point cloud and the manipulated object point cloud. 
    At time step $t$, ForeFormer takes as input the background point cloud with embedded background object names $P_{b} \in \mathbb{R} ^ {N^b \times (3+d_n)}$, object point cloud with embedded object name $P_{o} \in \mathbb{R} ^ {N^o \times (3+d_n)}$, task instructions $C$, and the noisy object goal point cloud $P_{o'}^{t}\in \mathbb{R} ^ {N^o \times 3}$, and outputs point-wise noise corresponding to the input noisy object goal point cloud.
    $N^b$ and $N^o$ represent the number of points in the background point cloud and the object point cloud, respectively.
    $d_n$ is the embedding dimension for the names of the background objects and the object to be manipulated.
    ForeFormer formulation is defined as:
    \begin{align}
		P_{o'}^0 \sim p_{\theta} (P_{o'}^0 ~|~ P_b, P_o, \phi(C)),
        \\
        where ~~
        p_{\theta} (P_{o'}^0 ~|~ P_b, P_o, \phi(C)) \approx \prod_{t=1}^{T} p_{\theta} (P_{o'}^{t-1} ~|~ P_{o'}^{t}, P_b, P_o, \phi(C)).
		\label{eq_ForeFormer}
	\end{align}
    where $p_{\theta}$ denotes the parameterized conditional probability distribution predicted by the ForeFormer network with weights $\theta$.
    $P_{o'}^{T} \sim \mathcal{N}(0,I)$ denotes standard Gaussian noise, $T$ is the total number of diffusion time steps, and $\phi(\cdot)$ represents the text encoder.
    For a more detailed description of CDDPMs, please refer to \cite{ho2020denoising}.

    To accurately predict point-wise noise, ForeFormer is built upon a Transformer architecture.
    It primarily consists of four modules: 1) object point cloud encoder, 2) background point cloud encoder, 3) feature fusion backbone, and 4) point cloud noise decoder.
    The overall structure of ForeFormer is shown in Figure.\ref{fig_ForeFormer} (a).
    A detailed description of each module is provided below.

    \textbf{Feature fusion backbone}
    We first introduce the feature fusion backbone, as it determines how ForeFormer encodes each input.
    To output noise that corresponds point-wise to the object point cloud, the feature fusion backbone employs a cross-attention mechanism \cite{vaswani2017attention}: the point-wise features of the object point cloud serve as query vectors, while the point-wise features of the background point cloud, concatenated with the diffusion time step and text features, serve as key and value vectors.
    The diffusion time step $t$ is represented via sinusoidal positional encoding and further mapped into the feature space through a multi-layer perceptron (MLP).
    Task instructions are encoded using the CLIP model \cite{radford2021learning} and then dimensionally aligned through an MLP.
    Detailed model parameters are provided in the Appendix.
    
    \textbf{Object point cloud encoder}
    ForeFormer first leverages the CLIP text encoder \cite{radford2021learning} to extract the embedding of the object name, replicates it to match the number of points, and fuses it into the object point cloud. The fused point cloud features are then modeled through a self-attention mechanism.
    Subsequently, an MLP expands the dimensionality of the object point cloud containing only coordinates, which is added to the point cloud features, thereby preserving point-wise positional details while effectively enhancing the contextual representation of the object.
    Next, the noisy object target point cloud is aligned in dimension with the object point cloud features via an MLP, summed, and fed into the feature fusion backbone for further processing.
    
    \textbf{Background point cloud encoder}
    The background point cloud consists of the task-relevant objects together with the CLIP embeddings of their corresponding names.
    Dense and excessive background point clouds can dilute attention during feature fusion, making it difficult to highlight context information truly relevant to the manipulated object and thereby degrading model performance.
    To address this, ForeFormer employs PointTransformer-v3 \cite{wu2024point} to encode the background point clouds. 
    PointTransformer-v3 leverages sparse convolutions and serialized pooling to gradually reduce point cloud density while effectively preserving both local and global contextual features. 
    Similarly, the background point cloud containing only coordinates is processed by an MLP and then added to the features extracted by PointTransformer-v3.
    Consequently, ForeFormer is able to extract more compact and informative background point cloud representations.
    
    \textbf{Point cloud noise decoder}
    ForeFormer employs PointNet \cite{qi2017pointnet} to decode the output of the feature fusion backbone, enabling per-point feature reconstruction of the object point cloud and thereby predicting the corresponding diffusion noise for each point.

    \subsection{Structural Consistency Loss}
    Typically, generative models based on CDDPMs are optimized by minimizing the error between predicted noise and true noise, ensuring the accuracy and diversity of the generated data. 
    However, for 3D goal state generation tasks, maintaining the structural consistency of objects is critical, meaning the spatial relationships between different points on the object surface should remain invariant before and after transformation.
    To address this, we have designed a structural consistency loss function, which is formulated as follows:
    \begin{align}
		\nonumber
        & \mathcal{L}_{Diff} = \\ 
        \nonumber
        &\mathbb{E}_{ P_{o'}^{*}, \epsilon \sim \mathcal{N}(0,I), t } \left[ \left \| \epsilon-\epsilon_{\theta}(P_{o'}^{t}, t, P_b, P_o, \phi(C))  \right \|_2^2 \right] \\ 
        &+ \omega \cdot \mathbb{E}_{ i,j \in \{1,2,...,N^o\} } \left[ (\left \| P_o[i] - P_o[j]  \right \|_2^2 - \left \| P_{o'}^0[i] - P_{o'}^0[j]  \right \|_2^2  ) ^2 \right]
		\label{eq_loss}
	\end{align}
    The first part of the equation calculates the mean squared error (MSE) between the predicted diffusion noise $\epsilon_{\theta}$ and the true noise $\epsilon$, while the second part computes the MSE between the self-distance matrices of the generated object point cloud in the goal state $P_{o'}^0$ and the input object point cloud $P_o$. A smaller error of self-distance matrices indicates greater structural consistency in the point cloud.
    It is important to note that when computing the self-distance matrix of the point cloud, the input includes only the point cloud coordinates and does not include the embedding of the object name.
    $\omega$ represents the weight of the structural consistency loss.
    $P_{o'}^0$ is the noise-free target state point cloud obtained by applying reverse diffusion to $P_{o'}^{t}$ using ForeFormer.
    To accelerate the training process, we use one-shot denoising instead of iterative denoising, as shown by the following formula:
    \begin{align}
		P_{o'}^0 = \frac{1}{\sqrt{\bar{\alpha}^k}}(P_{o'}^t - \sqrt{1-\bar{\alpha}^k} \cdot \epsilon_\theta(P_{o'}^{t}, t, P_b, P_o, \phi(C))
		\label{eq_}
	\end{align}
    where $\alpha^t$ is the noise scheduling parameter at timestep $t$, and $\bar{\alpha}^t=\begin{matrix} \prod_{t=1}^T \alpha_t \end{matrix}$.

    \subsection{Robot Motion Planning}
    Leveraging the point-wise correspondences predicted by ForeFormer, we efficiently estimate the rigid transformation of the object via Singular Value Decomposition (SVD), which significantly simplifies motion planning for the manipulator. 
    During execution, we combine grasp and push modes within a synchronous closed-loop planning framework.  
    If the object is not grasped, AnyGrasp \cite{fang2023anygrasp} is employed to generate candidate grasp poses, from which we select those that enable collision-free grasping and subsequent motion toward the target pose.
    For tasks without feasible grasps (e.g., ``pressing the key on a keyboard"), a push strategy guided by the point cloud displacement direction is adopted to move the object toward the goal. 
    The pipeline of robot motion planning is shown in Figure.~\ref{fig_ForeFormer} (b).
    The calculation method for the push starting point, direction, and distance is detailed in the Appendix.

    We observe that in most object manipulation tasks, rotations typically occur after translations. 
    For instance, in the task of ``Lift the kettle above the cup and tilt it to pour water", the kettle must first be translated above the cup before tilting to avoid spilling. 
    Existing approaches such as Robotwin \cite{mu2024robotwin} \cite{chen2025robotwin} leverage the encoding capability of large language models (LLMs) to generate constraint-satisfying robot motion programs, but they rely on detecting the object's affordance regions. 
    To address this, we propose a translation-prioritized motion planning method that enforces rotations to lag behind translations by decomposing the trajectory. 
    After an object is grasped, we first generate an end-effector trajectory from the current pose to the target pose, where the target pose is determined by the object's rigid transformation. 
    We then iteratively sample positions backward along this trajectory and attempt to plan collision-free motions that only involve translation while keeping the orientation fixed, until a feasible solution is found. 
    The identified position is taken as the trajectory segmentation point, enabling the manipulator to first translate to this point and then move toward the target pose, thereby achieving a translation-prioritized planning strategy.

    \section{Simulation Experiments}
    \label{Simulation_Experiments}
    We design a comprehensive set of experiments to thoroughly evaluate the performance of ForeFormer. The evaluation covers both simulated and real environments, seen and unseen scenes, over 20 tasks, multiple manipulation modes, and dozens of rigid and articulated objects. Through these experiments, we aim to answer the following questions:
    (1) Is the structural design and optimization of the ForeFormer network reasonable?
    (2) Does ForeFormer outperform previous 3D goal state prediction models?
    (3) Can ForeRobo achieve zero-shot sim-to-real?
    (4) What are the boundaries of ForeFormer's generalization capability?
    
    This section first presents the experiments conducted in the simulation environment.
    We find ForeFormer to significantly outperform the prior state-of-the-art on almost all of the tested simulation tasks, with an average success-rate improvement of 75.75\%.
    In the following sections, we provide an overview of each task, our evaluation methodology on that task, and our key takeaways.

    \begin{figure*}[tp]
		\centering
		{\includegraphics[scale=1]{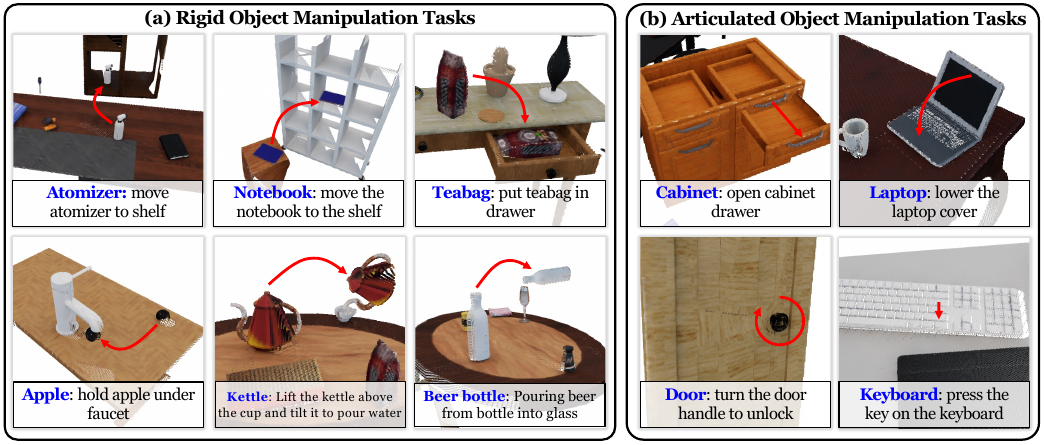}}
		\caption{
			\textbf{Simulation tasks}.
            We evaluated the performance of ForeFormer and the baseline methods on ten simulated tasks, including six rigid-object manipulation tasks and four articulated-object manipulation tasks.
            In the captions below each task illustration, the manipulated object is highlighted in \textcolor[rgb]{0,0,1}{blue}, followed by the task description.
            The red arrows indicate the transformation of objects from their initial states to the goal states that satisfy the task description.
		}
		\label{fig_simtasks}
	\end{figure*}

    \begin{figure*}[tp]
		\centering
		{\includegraphics[scale=0.9]{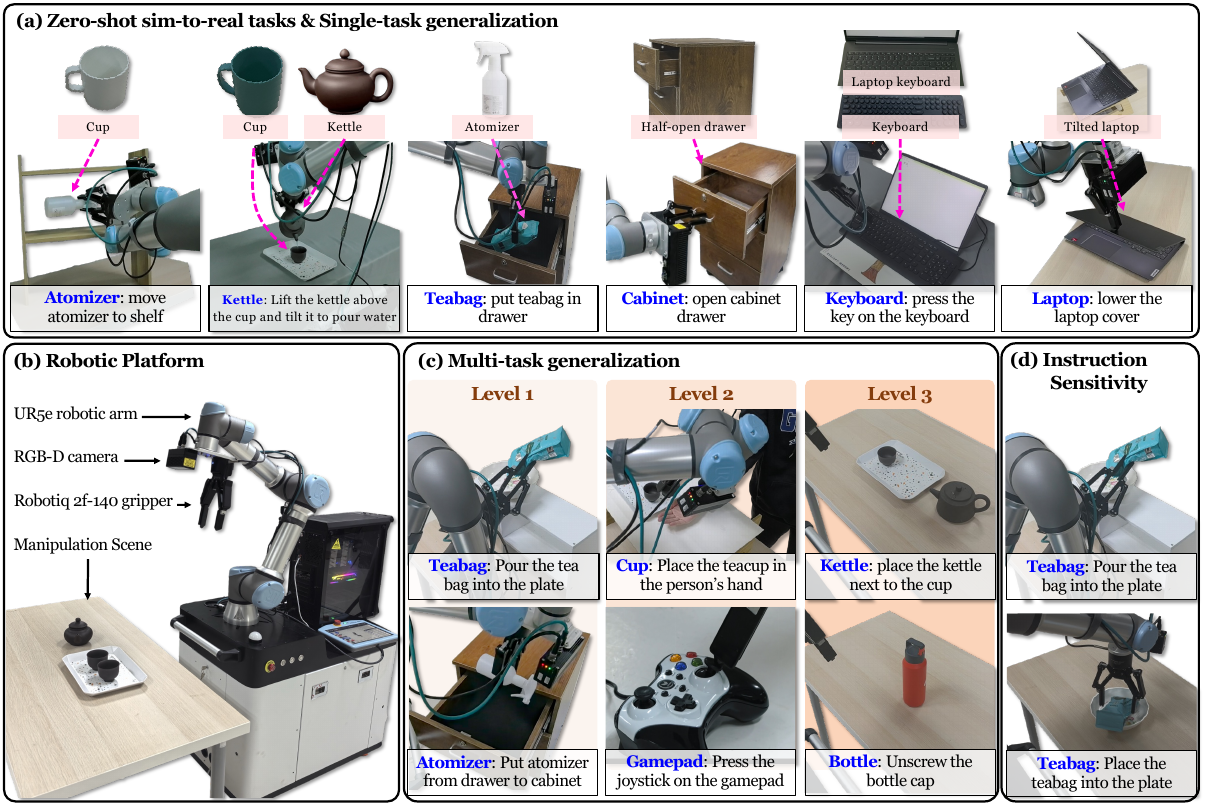}}
		\caption{
			\textbf{Real-world tasks and robotic platform}.
            Based on the zero-shot sim-to-real experiments, we constructed single-task generalization experiments by replacing the objects in each task.
            In each generalization experiment, only one object was replaced.
		}
		\label{fig_realtasks}
	\end{figure*}
    
    \subsection{Evaluation Tasks}
    We conduct experiments on six rigid-body manipulation tasks and four articulated-object manipulation tasks selected from the ForeMani-v1 dataset, which was constructed based on ForeGen, as shown in Fig.~\ref{fig_simtasks}.
    This work primarily focuses on single-object interaction tasks; thus, each task scene contains only one background object and one manipulated object.
    These tasks cover a wide range of object motion scales, diverse manipulation modes, goal states with or without contact, and either unimodal or multimodal prediction distributions.
    We evaluate different methods using either the point cloud error (PCE) of the predicted goal state or the success rate (SR) from simulation-based validation.
    Specifically, the point cloud error is defined as the mean minimum point-to-point distance between the predicted and ground-truth goal point clouds, and it is used for all articulated objects as well as the apple, kettle, and beer bottle tasks.
    The simulation-based validation directly sets the object transformation predicted by the model in the simulator and automatically determines correctness according to the task conditions, which is applied to the atomizer, notebook, and teabag tasks.
    Each task consists of 400 training samples and 400 testing samples.
    The training set covers 20 scenes, with 20 samples per scene.
    The testing set includes 20 seen scenes and 20 unseen scenes, with 10 samples per scene.
    Samples within the same scene share the same object model but differ in object placement.

	\begin{table}[tp]
		\footnotesize 
		\begin{center}
			\resizebox{\linewidth}{!}{
            \begin{tabular}{l|c|ccc}
                \toprule[1pt]

                \multirow{2}*{Encoders} & Atomizer & Kettle & Drawer & Laptop \\
                & SR $\uparrow$ & \multicolumn{3}{c}{PCE $\downarrow$} \\

                \midrule[0.7pt]

                SAT & 0.600 / 0.545 & 0.1268 / 0.1254 & 0.0347 / 0.0432 & 0.0133 / 0.0118 \\
                PTv3 & \textbf{0.820 / 0.715} & \textbf{0.0163 / 0.0312} & \textbf{0.0153 / 0.0218} & \textbf{0.0051 / 0.0040} \\

                \midrule[0.7pt]

                Improvement & 22.00\% / 17.00\% & 87.15\% / 75.12\% & 55.91\% / 49.54\% & 61.65\% / 66.10\% \\
                
                \bottomrule[1pt]
            \end{tabular}
			}
		\end{center}
		\captionsetup{justification=justified, singlelinecheck=false}
		\caption{
            \textbf{Ablation study on background point cloud encoder}.
            We present results with different checkpoint selection methods in the format of (seen scenes) / (unseen scenes).
            $\uparrow$ indicates that higher SR corresponds to better performance, while $\downarrow$ indicates the opposite.
            Compared to using a self-attention Transformer (SAT), constructing the background point cloud encoder with PointTransformer-v3 significantly improves performance.
			}
		\label{tab_ablation_ptv3}
	\end{table}

    \begin{figure}[tp]
		\centering
		{\includegraphics[scale=0.44]{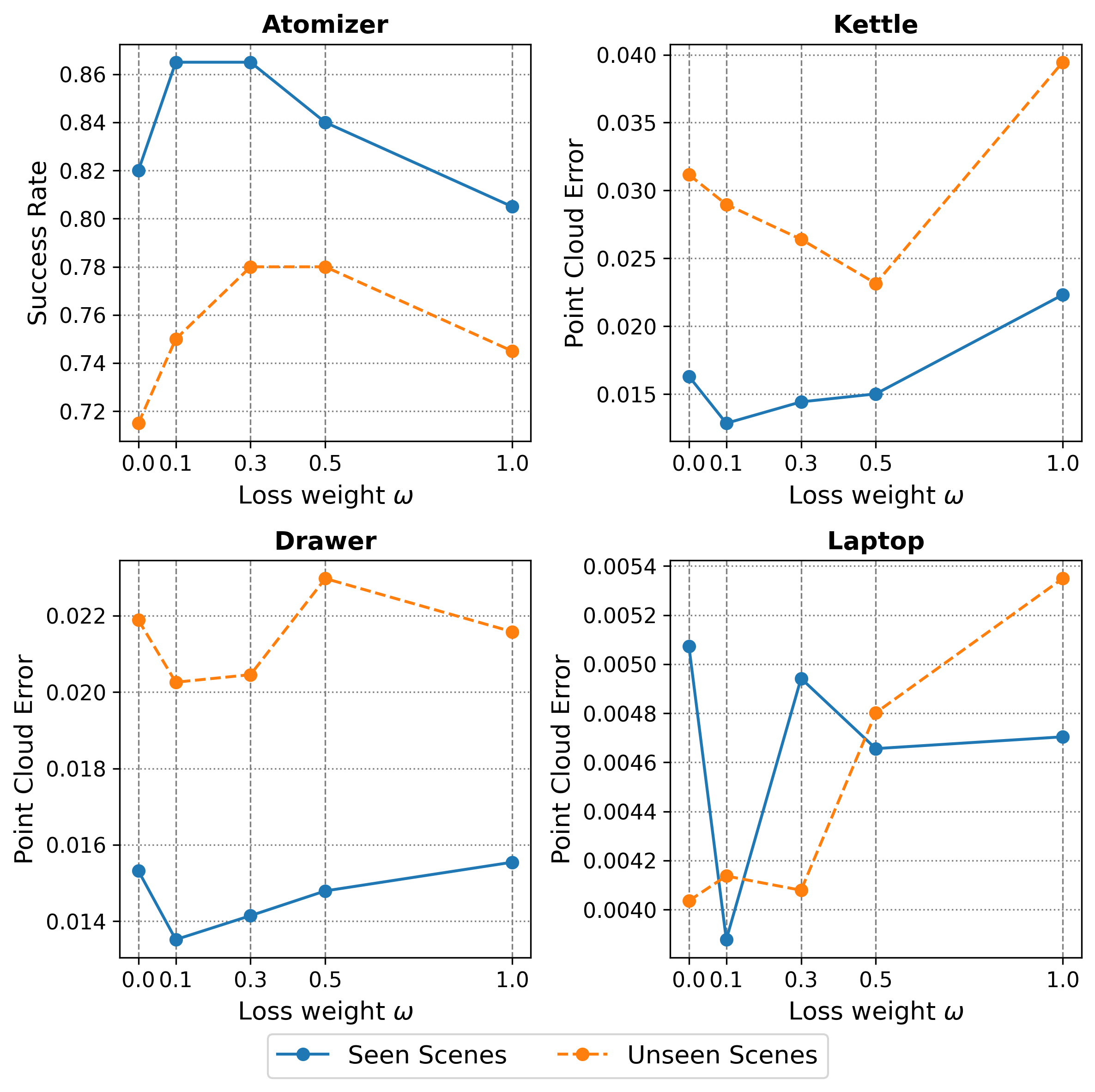}}
		\caption{
			\textbf{Ablation study on structural consistency loss weight}.
            The figure shows the performance of ForeFormer in both seen and unseen scenes for each task as the structural consistency loss weight increases. 
            For each task, we tested five different loss weights ranging from 0 to 1.
		}
		\label{fig_ablation_loss}
	\end{figure}

    \subsection{Ablation study}

    In this section, we perform several tests to validate the effectiveness of the ForeFormer network architecture and weight optimization.
    The evaluation includes two rigid-object manipulation tasks (atomizer and kettle) and two articulated-object manipulation tasks (drawer and laptop), covering four manipulation modes: diverse placement, suspended manipulation, translation, and rotation.
    We train a separate ForeFormer weight for each test task and comparison case.
    The experimental results are shown in Table.~\ref{tab_ablation_ptv3} and Figure.~\ref{fig_ablation_loss}.
    Except for the case where SAT is used as the background point cloud encoder, the other ForeFormer weights contain approximately 40M parameters. 
    Detailed model hyperparameters and training methods are provided in the Appendix.
    Following paragraphs summarize the key takeaways.
    
    \textit{A more compact background point cloud embedding significantly improves performance}:
    ForeFormer uses PTv3 to obtain embeddings of the background object point cloud. PTv3 generates more compact point cloud representations through a voxel-based serialization and pooling process, achieving approximately a $4\times$ compression ratio compared to the self-attention Transformer under our parameter settings.
    The results in Table.~\ref{tab_ablation_ptv3} show that PTv3 achieves an average performance improvement of 54.31\%, and also indicate that excessively dense background point clouds disperse attention during feature fusion, making it difficult to emphasize context relevant to the manipulated object, thereby reducing model performance.
    An interesting observation is that compact point cloud embeddings yield substantially greater performance gains on small-scale background point clouds than on large-scale ones. 
    For instance, in the Kettle task, the background object (a cup) has the smallest scale and shows the largest improvement (81.14\%), whereas in the Atomizer task, the background shelf has the largest scale and exhibits the smallest improvement (19.50\%). 
    This is because point cloud information in small-scale objects is more concentrated, enabling the pooling module to more effectively highlight key geometric features, and thus leading to more significant performance gains in tasks involving small-scale background objects than in those with large-scale backgrounds.
    
    \textit{ForeFormer generates structurally consistent point clouds}:
    ForeFormer introduces structural consistency loss to constrain the geometric structure consistency between the generated point cloud and the initial point cloud, enhancing local feature alignment and reducing unnecessary point cloud deformation and jitter, thereby improving the accuracy of object pose transformation calculated through SVD and ultimately enhancing prediction accuracy.
    As shown in Figure.~\ref{fig_ablation_loss}, compared to not using structural consistency supervision, a weight of 0.1 for the structural consistency loss resulted in an average performance improvement of 9.56\%.
    However, as the supervision weight increases, the model focuses excessively on local structures, suppressing generative diversity and consequently degrading performance. 
    Moreover, ForeFormer performs noticeably better in seen scenes than in unseen ones.

    \subsection{Evaluation Methodology}
    We compare ForeFormer with existing methods and test it on all 10 simulation tasks.
    For baseline methods, we compare with the cross 3D transformation estimation-based method (Taxpose \cite{pan2023tax}), iterative pose denoising-based method (RPDiff \cite{simeonov2023shelving}), paired point cloud generation-based method (Imagination Policy \cite{huang2025imagination}).
    These methods take 3D point clouds as the input format and are applied to tasks involving one manipulated object and one background object.
    RPDiff and Imagination Policy are designed specifically for pick-place tasks, where object point clouds are decentralized before being fed into the network, which disrupts the spatial relationships of articulated objects. 
    Therefore, for articulated object manipulation tasks, we removed the point cloud decentralization step from RPDiff and Imagination Policy.
    ForeFormer is trained for 6,000 epochs across all tasks and uses DDIM \cite{song2020denoising} during inference to accelerate prediction.
    The results from simulation benchmarks are summarized in Tables.~\ref{tab_sim_compare1} and \ref{tab_sim_compare2}.

	\begin{table*}[tp]
		\footnotesize 
		\begin{center}
			\resizebox{\linewidth}{!}{
            \begin{tabular}{l|ccc|ccc|c}
                \toprule[1pt]

                \multirow{2}*{Methods} & Atomizer & Notebook & Teabag & Apple & Beer bottle & Kettle & \multirow{2}*{Inference FPS}  \\
                
                & \multicolumn{3}{c}{SR $\uparrow$} & \multicolumn{3}{c}{PCE $\downarrow$} & \\

                \midrule[0.7pt]

                Imagination Policy & 0.45 / 0.59 & 0.555 / 0.355 & 0.0 / 0.005 & 0.1139 / 0.1361 & \textbf{0.0593 / 0.0751} & 0.8857 / 0.8857 & 0.04 \\
                
                RPDiff & \underline{0.775} / \underline{0.75} & \underline{0.695} / \underline{0.795} & 0.43 / 0.38 & 0.0915 / 0.1033 & 0.0876 / 0.0895 & 0.0724 / 0.0848 & 2.85 \\

                Taxpose & 0.58 / 0.605 & 0.525 / 0.535 & \underline{0.55} / \underline{0.4} & \underline{0.0299} / \underline{0.0798} & 0.1024 / 0.0900 & \underline{0.0258} / \underline{0.0334} & 74.49 \\
                					
                \midrule[0.7pt]

                \textbf{ForeFormer (Ours)} & \textbf{0.865 / 0.75} & \textbf{0.88 / 0.81} & \textbf{0.69 / 0.46} & \textbf{0.0187 / 0.0508} & \underline{0.0652} / \underline{0.0886} & \textbf{0.0128 / 0.0289} & 3.57 \\ 
                					
                \midrule[0.7pt]

                Improvement (vs. Taxpose) & 28.50\% / 14.50\%  & 35.50\% / 27.50\%  & 14.00\% / 6.00\% & 37.46\% / 36.34\% & 36.33\% / 1.55\% & 50.39\% / 13.47\% &  \\
                
                Improvement (vs. Best) & 9.00\% / 0.00\%  & 18.50\% / 1.50\%  & 14.00\% / 6.00\% & 37.46\% / 36.34\% & -9.94\% / -17.98\% & 50.39\% / 13.47\% &  \\
                
                \bottomrule[1pt]
            \end{tabular}
			}
		\end{center}
		\captionsetup{justification=justified, singlelinecheck=false}
		\caption{\textbf{Comparison in simulated rigid object manipulation tasks}.
            Performance is reported in the same format as in Tab.~\ref{tab_ablation_ptv3}.
            The best results under the same comparison condition are highlighted in \textbf{bold}, while the second-best results are \underline{underlined}.
            ForeFormer achieves an average improvement of 25.13\% and 13.23\% compared to the results of the optimal baseline method (Taxpose) and the best results of all baseline methods, respectively.
            The calculation formula for performance improvement is shown in the Appendix.
			}
		\label{tab_sim_compare1}
	\end{table*} 

	\begin{table*}[tp]
		\footnotesize 
		\begin{center}
            \begin{tabular}{l|cccc}
                \toprule[1pt]

                \multirow{2}*{Methods} & Door &	Cabinet & Keyboard & Laptop  \\
                
                & \multicolumn{4}{c}{PCE $\downarrow$} \\

                \midrule[0.7pt]

                Imagination Policy & 0.4299 / 0.3632 & 0.8857 / 0.8287 & \underline{0.0684} / \underline{0.0601} & 0.1476 / 0.1419 \\
                
                RPDiff & \underline{0.2223} / \underline{0.2194} & 7.0391 / 7.0188 & 0.2278 / 0.2241 & 0.5021 / 0.5231 \\
                			
                Taxpose & 0.3052 / 0.3144 & \underline{0.0271} / \underline{0.0279} & 0.1132 / 0.1251 & \underline{0.0276} / \underline{0.0341} \\
                					
                \midrule[0.7pt]

                \textbf{ForeFormer (Ours)} & \textbf{0.0089 / 0.0167} & \textbf{0.0135 / 0.0202} & \textbf{0.0011 / 0.0019} & \textbf{0.0038 / 0.0041} \\
                					
                \midrule[0.7pt]

                Improvement (vs. Taxpose) & 97.08\% / 94.68\% & 50.18\% / 27.59\% & 99.02\% / 98.48\% & 86.23\% / 87.97\% \\
                
                Improvement (vs. Best) & 95.99\% / 92.38\% & 50.18\% / 27.59\% & 98.39\% / 96.83\% & 86.23\% / 87.97\% \\
                
                \bottomrule[1pt]
            \end{tabular}
		\end{center}
		\captionsetup{justification=justified, singlelinecheck=false}
		\caption{
            \textbf{Comparison in simulated articulated object manipulation tasks}.
            Performance is reported in the same format as in Tab.~\ref{tab_sim_compare1}.
            ForeFormer achieves an average improvement of 80.15\% and 79.45\% compared to the results of the optimal baseline method (Taxpose) and the best results of all baseline methods, respectively.
			}
		\label{tab_sim_compare2}
	\end{table*}

    \subsection{Key Findings}
    ForeFormer outperforms alternative methods on all tasks and variants in our simulation tasks study (Tables.~\ref{tab_sim_compare1} and \ref{tab_sim_compare2}) with an average improvement of 47.14\% and 39.72\% compared to the results of the optimal baseline method (Taxpose) and the best results of all baseline methods, respectively.
    Following paragraphs summarize the key takeaways.
    
    \textit{ForeFormer is capable of handling diverse manipulation modes}:
    ForeFormer achieves the best performance across all rigid and articulated object manipulation tasks. This advantage stems from its input preprocessing and network design, which preserve the spatial structure of various task types (e.g., transforming articulated objects as a whole), as well as its attention-based ability to process variable-length point clouds, enabling robustness to different object scales and point densities. These results suggest that ForeFormer has strong potential to extend to more complex manipulation domains, such as deformable and flexible objects.
    
\begin{figure}[tp]
		\centering
		{\includegraphics[scale=0.35]{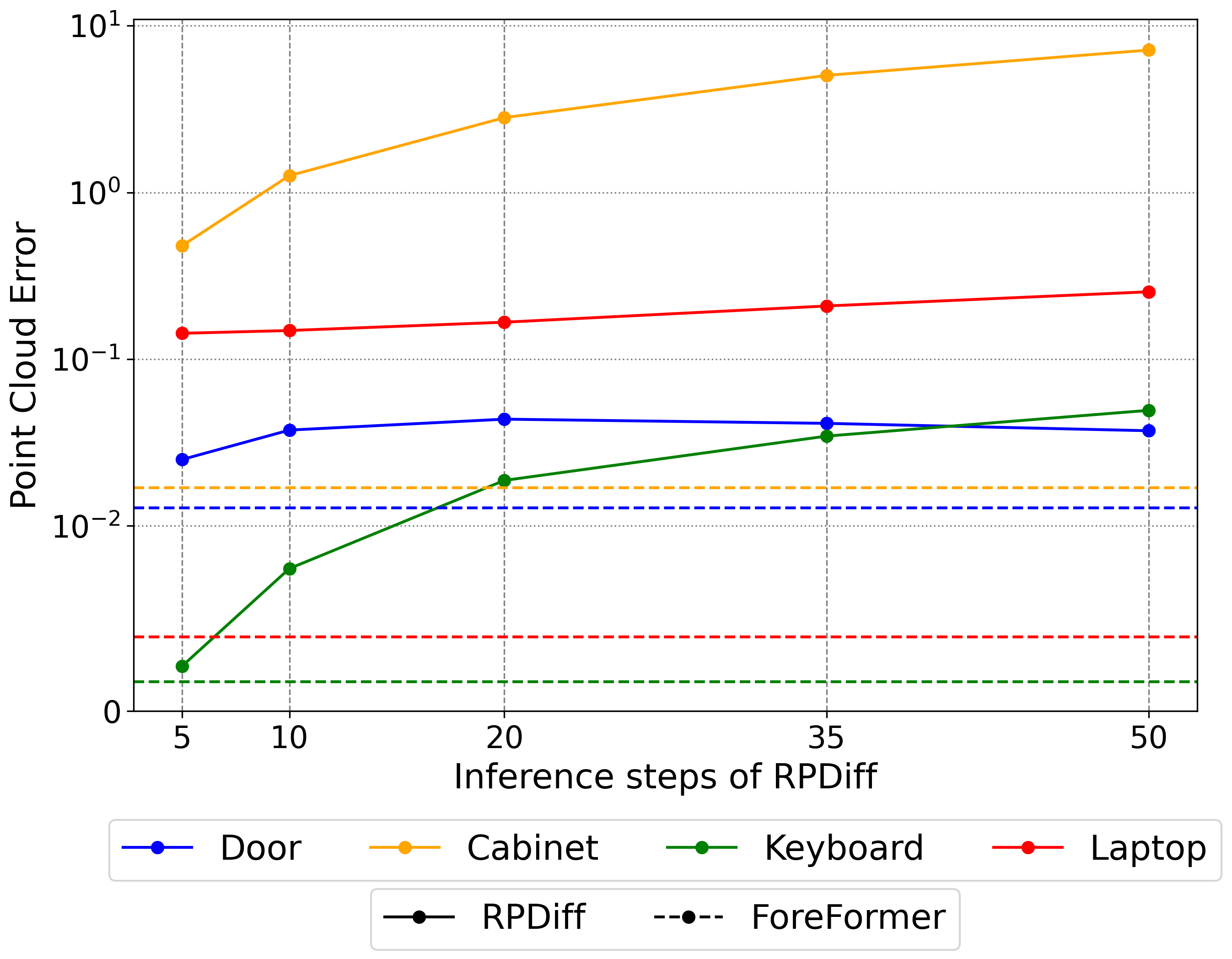}}
		\caption{
			\textbf{Comparison Between ForeFormer and RPDiff with Different Inference Steps}.
            The solid lines represent the point cloud error of RPDiff, while the dashed lines correspond to ForeFormer. 
            The values 5 and 50 denote the training and inference steps reported by RPDiff, respectively. As the number of inference steps increases, the error of RPDiff continues to grow, with this trend being most pronounced for prismatic articulated objects (e.g., cabinet and keyboard). 
            Overall, ForeFormer consistently outperforms RPDiff across all scenarios.
		}
		\label{fig_rpdiff}
	\end{figure}

    \textit{Each baseline method struggles with certain types of cases}:
    Taxpose estimates cross-object poses by predicting the importance weights of points between the operating and background objects. This makes small objects more susceptible to estimation errors of larger objects, resulting in degraded performance when object scales differ significantly (e.g., teabag, door, and keyboard tasks).
    RPDiff learns uniformly distributed transformation trajectories from the initial to target poses (referred to as "denoising" in their paper). However, for articulated objects, the initial states and motion patterns are too uniform, resulting in target states that lack consistent structural constraints. Consequently, during denoising, the model can only accumulate joint increments step by step without converging to a structurally consistent target state, leading to increased errors as the number of inference steps grows (see Figure.~\ref{fig_rpdiff}).
    The Imagination Policy performs poorly on most tasks, which may be attributed to the limited expressive capacity of the model.

    \textit{ForeFormer exhibits a minor generation mode divergence}:
    This manifests as a single generation of point clouds clustering into multiple object contours at different positions, each corresponding to a feasible target state.
    We employ an iterative rigid point cloud registration algorithm with a Huber robust kernel to select the point cloud cluster with the highest structural consistency from the predicted point clouds and compute its transformation relative to the initial object point cloud, thereby mitigating the impact of generation mode divergence.
    This phenomenon is illustrated in the fifth subfigure of the Kettle task in Figure.~\ref{fig_real_denoising}, where the goal point cloud of the kettle is clustered on both sides above the cup. 
    The rightmost subfigure shows the result after applying the iterative rigid point cloud registration algorithm, where the goal point cloud is successfully consolidated into the desired goal state on the left side.

    \section{Real-world Evaluation}
    We evaluated ForeFormer and the baseline methods on 21 real-world tasks, with all methods trained solely on simulation data generated by ForeGen.
    Across all comparisons, ForeFormer achieves an average improvement of 65.50\% over the best-performing baseline, Taxpose \cite{pan2023tax}.
    More details about the real world setup and parameters may be found in supplemental materials.

    \begin{table*}[tp]
		\footnotesize 
		\begin{center}
			\resizebox{\linewidth}{!}{
            \begin{tabular}{l|c@{\hskip 4pt}c@{\hskip 4pt}|c@{\hskip 4pt}c@{\hskip 4pt}c@{\hskip 4pt}|c@{\hskip 4pt}c@{\hskip 4pt}|c@{\hskip 4pt}c@{\hskip 4pt}|c@{\hskip 4pt}c@{\hskip 4pt}c@{\hskip 4pt}|c@{\hskip 4pt}c}
                \toprule[1pt]
                
                \multirow{2}*{Methods} & \multicolumn{2}{c|}{Atomizer} & \multicolumn{3}{c|}{Kettle} & \multicolumn{2}{c|}{Teabag} & \multicolumn{2}{c|}{Cabinet} & \multicolumn{3}{c|}{Keyboard} & \multicolumn{2}{c}{Laptop}  \\

                & Bas. & Cup & Bas. & Cup & Kettle & Bas. & Atomizer & Bas. & Half-open drawer & Bas. & Laptop keyboard & Keyboard & Bas. & Tilted laptop \\
                
                \midrule[0.7pt]
                
                RPDiff & 0.10 & 0.0 & 0.0 & 0.05 & 0.10 & 0.30 & 0.40  & -- & -- & -- & -- & -- & -- & -- \\
                			
                Taxpose & 0.0 & 0.0 & 0.35 & 0.0 & 0.30 & 0.0 & 0.0 & 0.0 & 0.0 & 0.0 & 0.0 & 0.0 & 0.70 & 0.65 \\
                					
                \midrule[0.7pt]

                \textbf{ForeFormer (Ours)} & \textbf{0.80} & \textbf{0.70} & \textbf{0.75} & \textbf{0.60} & \textbf{0.70} & \textbf{0.60} & \textbf{0.70} & \textbf{0.395} & \textbf{1.0} & \textbf{1.0} & \textbf{1.0} & \textbf{1.0} & \textbf{1.0} & \textbf{0.925} \\
                					
                \midrule[0.7pt]

                Improvement (vs. Best) & 70\% & 70\% & 40\% & 55\% & 40\% & 30\% & 30\% & 39.5\% & 100\% & 100\% & 100\% & 100\% & 30\% & 27.5\%\\
                
                \bottomrule[1pt]
            \end{tabular}
			}
		\end{center}
		\captionsetup{justification=justified, singlelinecheck=false}
		\caption{
            \textbf{Zero-shot sim-to-real and single-task generalization experimental results}.
            The results are reported in terms of success rate. 
            For each task, we present the basic sim-to-real experimental results, which are closely aligned with the simulated objects, as well as the generalization experiments involving object replacement. The replaced objects are shown in Figure.~\ref{fig_realtasks} (a).
            Each reported result corresponds to a model that is trained solely on its respective simulation task.
            Given that RPDiff exhibits large errors on articulated object manipulation tasks in simulation, we do not evaluate it in real-world settings to avoid potential safety risks.
			}
		\label{tab_sim2real}
	\end{table*}

    \subsection{Zero-shot Sim-to-real}  
    In this section, we evaluate ForeRobo's zero-shot sim-to-real transfer capability across six tasks, including three rigid object manipulation tasks (kettle/Atomizer/Teabag) involving non-contact interactions and 3D placement, as well as three articulated object manipulation tasks (Cabinet/Keyboard/Laptop) involving both prismatic and revolute joint operations.
    Figure.~\ref{fig_realtasks} (a) presents video snapshots of the robot executing the tasks in the real world.
    Our robotic platform consists of a UR5e manipulator, a Robotiq 2F-140 gripper, a supporting base, and a Percipio depth camera mounted at the end effector, which provides a depth accuracy of approximately 1-2 mm, as shown in Figure.~\ref{fig_realtasks} (b).
    All evaluated models take the instance-segmented object point cloud as input and predict a point-wise target point cloud along with the corresponding pose transformation. 
    The robot then performs motion planning based on the grasp pose on the grasping affordance region and the predicted object transformation. 
    The robot employs admittance control to handle minor collisions and contact disturbances.
    The overall algorithmic pipeline in real-world scenarios is illustrated in Figure.~\ref{fig_ForeRobo}, while the manipulator's motion planning process is shown in Figure.~\ref{fig_ForeFormer} (b). 
    Detailed experimental settings are provided in the Appendix.

    For each task, we conduct 10 repeated trials, randomly resetting the object's position within the robot's workspace before each trial. 
    For the Atomizer, Teabag, and Keyboard tasks, the success of a single trial is recorded as 1 or 0 based on whether the task description is fully achieved.
    For the Kettle task, a trial is assigned a success score of 1 if all the water is poured into the cup, 0.5 if some water is spilled but still partially poured into the cup, and 0 if none of the water reaches the cup.
    For the Cabinet task, the success rate is computed as the ratio between the drawer's pulled-out distance and its maximum pull-out distance. For the Laptop task, the success rate is defined as the ratio between the closed angle achieved and the initial open angle of the laptop.
    The results are shown in Table.~\ref{tab_sim2real}.

\begin{figure*}[tp]
		\centering
		{\includegraphics[scale=1.55]{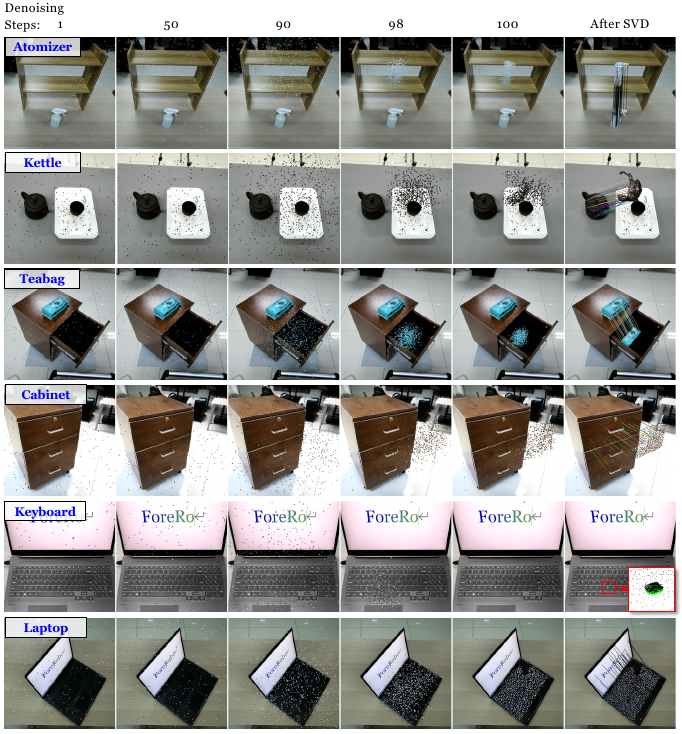}}
		\caption{
			\textbf{Illustration of ForeFormer's Inference Process in Real-World Scenes}.
            Each subfigure visualizes the point cloud produced by ForeFormer at the corresponding diffusion step, projected into the image space. 
            In the ``After SVD" column, the lines denote the point correspondences between the initial object point cloud and the predicted goal point cloud. 
            The inference process of ForeFormer consists of 100 denoising steps.
		}
		\label{fig_real_denoising}
	\end{figure*}

    ForeFormer achieves an average success rate of 75.75\%, representing a 51.58\% average improvement over the best resulrs of baseline methods.
    Moreover, in most tasks, its real-world performance closely matches its performance in simulation.
    A notable performance drop is observed only in the Cabinet task when transferring from simulation to the real world. This is primarily because the simulated training point clouds were captured from a relatively distant viewpoint, covering a large portion of the cabinet surface. 
    In contrast, the real-world camera is positioned closer to the object than the robot's typical operating distance, resulting in significantly fewer observable object features. 
    Consequently, ForeFormer occasionally predicts the drawer's target point cloud in the opposite direction.
    In addition, the typical failure cases of ForeFormer arise from large errors in the predicted target point cloud or unintended collisions with background objects, which may result in placing the object in an incorrect position or failing to generate a feasible motion plan.
    The inference process of ForeFormer in real-world tasks is shown in Figure.~\ref{fig_real_denoising}.

\begin{figure}[tp]
		\centering
		{\includegraphics[scale=1.3]{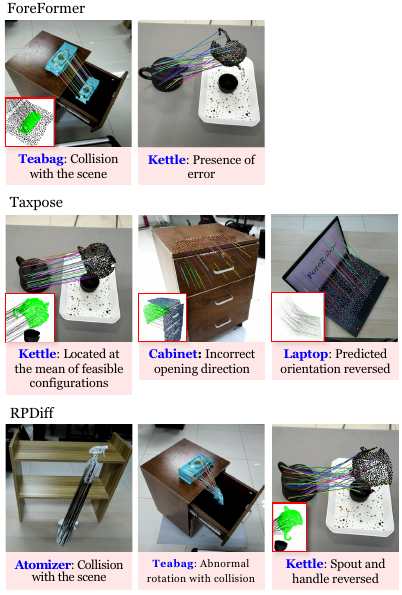}}
		\caption{
			\textbf{Representative failure cases of different methods}.
            For tasks that are difficult to distinguish, we additionally provide the corresponding 3D point cloud visualizations for clarification.
		}
		\label{fig_real_error}
	\end{figure}

    Compared with ForeFormer, TaxPose and RPDiff not only suffer from larger prediction errors and more frequent collisions with the environment, but also produce several unreasonable target states:
    (1) abnormal rotations that prevent the grasping algorithm from obtaining a valid grasp pose, for example, the predicted Teabag often tilts toward the unobserved side, causing all grasp poses detected in the initial state to collide in the predicted goal state;
    (2) goal states cokellapsing to the mean of multiple feasible configurations, such as predicting the kettle directly above the cup;
    (3) more pronounced inversion errors, including swapping the handle and spout of the kettle, reversing the drawer opening direction in the Cabinet task, and flipping the Laptop lid such that the hinge side and the opposite side are interchanged.
    Representative failure cases of different methods are shown in Figure.~\ref{fig_real_error}.

    \subsection{Single-Task Generalization Evaluation}  

    In this section, we evaluate the generalization capability of models trained on a single simulation task in real-world scenarios. 
    Specifically, building upon the zero-shot sim-to-real experiments, we replace the objects in the scene with others or different object states that were not seen during training, as shown in Figure.~\ref{fig_realtasks} (a). 
    The algorithm pipeline, testing steps, and evaluation metrics remain consistent with the zero-shot sim-to-real experiments. 
    The experimental results are presented in Table.~\ref{tab_sim2real}.

    ForeFormer achieves an average success rate of 82.81\% in the single-task generalization experiments, representing an 8.73\% improvement over the basic sim-to-real results and a 65.31\% improvement over the best-performing baseline. 
    In the Cabinet task, using a half-open drawer increases ForeFormer's success rate by 60.5\%, achieving zero failures. 
    This is because the half-open configuration provides prior information about the drawer's opening direction, preventing the model from predicting reversed motions or insufficient opening. 
    These results demonstrate that ForeFormer generalizes effectively to unseen objects and object states. 
    This capability stems from its point-cloud-level feature fusion, point-wise prediction paradigm, and sampling-based global point cloud denoising, which enable the model to learn local contact interaction patterns between the manipulated object and the background object rather than relying solely on global object position. 
    Consequently, these local interaction patterns can be transferred to unseen point cloud distributions.
    In contrast, the baseline methods still perform poorly in these scenarios and fail to generalize to unseen objects, such as the new cup in the Kettle task.

	\begin{table}[tp]
		\footnotesize 
		\begin{center}
			\resizebox{\linewidth}{!}{
            \begin{tabular}{cc|cc|cc}
                \toprule[1pt]

                \multicolumn{2}{c}{Level 1} & \multicolumn{2}{c}{Level 2} & \multicolumn{2}{c}{Level 3} \\

                Teabag & Atomizer & Cup & Gamepad & Kettle & Bottle \\

                \midrule[0.7pt]

                0.6 & 0.8 & 0.6 & 1.0 & 0.0 & 0.0 \\
                
                \bottomrule[1pt]
            \end{tabular}
			}
		\end{center}
		\captionsetup{justification=justified, singlelinecheck=false}
		\caption{
            \textbf{Results of ForeFormer in the multi-task generalization experiments.}
			}
		\label{tab_multi_task}
	\end{table} 

    \subsection{Multi-Task Generalization Evaluation}   

    In this section, we evaluate the generalization limits of ForeFormer trained under a multi-task setting. 
    We perform joint training on the ForeMani-v1 dataset constructed using the proposed ForeGen method, which consists of 106 distinct tasks and a total of 41,810 training samples. 
    ForeFormer is scaled to approximately 110M parameters to accommodate the diverse task set and enhance its overall generalization capability; detailed hyperparameters are provided in the Appendix. 
    The training process is conducted on eight H100 GPUs and takes approximately 20 days to complete.


    The evaluation is divided into three levels: \textit{Level 1} uses seen interaction patterns with seen object combinations, \textit{Level 2} uses seen interaction patterns with unseen object combinations, and \textit{Level 3} uses unseen interaction patterns with seen object combinations. 
    Here, an ``interaction pattern" refers to the goal contact or interaction behavior between the manipulated object and the background object, such as placing object A on or inside object B, or pouring the contents of object A into object B. 
    These levels impose progressively stronger generalization requirements on the model. 
    Each level includes two testing tasks, as illustrated in Figure.~\ref{fig_realtasks} (c).
    Except for conducting five trials per task, the algorithmic pipeline, testing procedure, and evaluation metrics remain identical to those used in the zero-shot sim-to-real experiments.
    The experimental results are presented in Table.~\ref{tab_multi_task}.
    
    ForeFormer achieves an average success rate of 75\% on Level 1 and Level 2 tasks, demonstrating strong generalization to seen interaction patterns with different object combinations. 
    In contrast, it achieves a 0\% success rate on Level 3 tasks, indicating an inability to generalize to unseen interaction patterns, consistent with previous manipulation strategies. 
    This limitation could potentially be mitigated by incorporating training data with diverse interaction patterns or by employing a 3D-aligned task text embedding model.

    Additionally, we evaluated the text instruction sensitivity of ForeFormer. 
    Specifically, in manipulation scenes containing the same objects, we tested whether the model could generate the corresponding target state when given different task descriptions. 
    The test scene is shown in Figure.~\ref{fig_realtasks} (d), with the manipulated object being a teabag and the background object a plate. The instructions used for testing were ``pour the teabag into the plate" and ``place teabag into the plate," differing by only a single word. 
    Each task was tested five times, yielding success rates of 80\% and 60\%, respectively. 
    The single failure in the ``pour teabag" task was due to the robotic arm being unable to plan a path, although the predicted target state largely matched the task description. 
    The ``place teabag" task had more failures; when the teabag stood upright on the table, the model occasionally predicted the target state corresponding to ``pour teabag," indicating that ForeFormer's predictions are influenced by the object's initial state. 
    This behavior can be attributed to the training data, which included only the task ``pour the teabag into the plate" for teabag and plate, but not ``place teabag." 
    Overall, the results indicate that ForeFormer exhibits a certain level of sensitivity to textual instructions.

    \section{Disscution}
    The primary goal of ForeRobo is to explore the significance and effectiveness of simulation data in enhancing real-world robotic manipulation.
    It consists of two core components:
    (1) ForeGen, the first system capable of generating unlimited high-precision manipulation states in simulation from a single human demonstration; and
    (2) ForeFormer, the first model that enables zero-shot sim-to-real transfer for 3D goal state generation.
    Through extensive experiments, we address the four key questions proposed in Section.~\ref{Simulation_Experiments}:
    (1) The network design and optimization of ForeFormer are well-structured and effective, achieving a performance improvement of approximately 31.94\% over alternative methods.
    (2) ForeFormer consistently outperforms prior 3D state prediction models, with an average improvement of 56.32\% compared to the best baseline.    
    (3) ForeRobo achieves zero-shot sim-to-real transfer, reaching an average success rate of 75.75\% in real-world scenarios.
    (4) ForeFormer demonstrates strong generalization across seen manipulation patterns and unseen object combinations, achieving an average success rate of 82.81\% in real-world generalization tests.
    In addition, although not presented in this paper, ForeFormer directly models fine-grained 3D states rather than object poses, making it naturally extendable to flexible and deformable objects. This will be an important direction of our future work.
    Overall, ForeRobo highlights the enduring potential and practical value of simulation data for robotic learning.

    Although ForeRobo has demonstrated effective learning and generalization capabilities in the selected tasks, it still has some limitations and room for improvement.
    First, the current version of ForeGen cannot yet handle flexible or deformable objects, as their states cannot be adequately represented using only object poses and joint angles. 
    Second, ForeFormer lacks explicit structural preservation, which may lead to noticeable deformation or jitter in the generated 3D states. 
    In future work, we will continue to improve ForeRobo to further enhance the applicability and robustness of simulation data in real-world scenarios.


	\bibliographystyle{Bibliography/IEEEtranTIE}
	\bibliography{Bibliography/IEEEabrv, Bibliography/reference}\


\end{document}